\documentclass[letterpaper]{article} % DO NOT CHANGE THIS
\usepackage{aaai2026}  % DO NOT CHANGE THIS
\usepackage{times}  % DO NOT CHANGE THIS
\usepackage{helvet}  % DO NOT CHANGE THIS
\usepackage{courier}  % DO NOT CHANGE THIS
\usepackage[hyphens]{url}  % DO NOT CHANGE THIS
\usepackage{graphicx} % DO NOT CHANGE THIS
\usepackage{csquotes}
\usepackage{float}
\usepackage{dblfloatfix} % 
\usepackage{svg}
\usepackage{booktabs}
\usepackage{multirow}
\usepackage{colortbl}
\usepackage[table]{xcolor}
\usepackage{array}
\usepackage{makecell}
\usepackage{natbib}  % DO NOT CHANGE THIS AND DO NOT ADD ANY OPTIONS TO IT
\usepackage{caption} % DO NOT CHANGE THIS AND DO NOT ADD ANY OPTIONS TO IT
\usepackage{algorithm}
\usepackage{algorithmic}
\usepackage{enumitem}  % enumerate
\usepackage{tikz}      % 

\newcommand*\circled[1]{\tikz[baseline=(char.base)]{
    \node[shape=circle,draw,inner sep=0.8pt] (char) {\scriptsize #1};}}

\usepackage{pifont}

\newcommand{\cmark}{\ding{51}} %  ✔
\newcommand{\xmark}{\ding{55}} %  ✘
\usepackage{newfloat}
\usepackage{listings}

\DeclareCaptionStyle{ruled}{labelfont=normalfont,labelsep=colon,strut=off} % DO NOT CHANGE THIS
\floatstyle{ruled}
\newfloat{listing}{tb}{lst}{}
\floatname{listing}{Listing}
\title{CreBench: Human-Aligned Creativity Evaluation from Idea to Process to Product}
\author{
    Kaiwen Xue\textsuperscript{\rm 1}\equalcontrib,Chenglong Li\textsuperscript{\rm 1}\equalcontrib,Zhonghong Ou\textsuperscript{\rm 2\ †},Guoxin Zhang\textsuperscript{\rm 1},Kaoyan Lu\textsuperscript{\rm 3},Shuai Lyu\textsuperscript{\rm 1},Yifan Zhu\textsuperscript{\rm 1},Ping Zong \textsuperscript{\rm 1},Junpeng Ding\textsuperscript{\rm 1},Xinyu Liu\textsuperscript{\rm 4},Qunlin Chen\textsuperscript{\rm 4},Weiwei Qin\textsuperscript{\rm 1},Yiran Shen\textsuperscript{\rm 1},Jiayi Cen\textsuperscript{\rm 5}
    \thanks{Corresponding authors}
}
\affiliations{
     \textsuperscript{\rm 1}Beijing University of Posts and Telecommunications\\
    \textsuperscript{\rm 2}State Key Laboratory of Networking and Switching Technology, Beijing University of Posts and Telecommunications\\
    \textsuperscript{\rm 3}School of Materials Science and Engineering, Shanghai Jiao Tong University\\
    \textsuperscript{\rm 4}Faculty of Psychology, Southwest University\\
    \textsuperscript{\rm 5}Southampton Education School, University of Southampton\\
    xkw@bupt.edu.cn, chenglong@bupt.edu.cn, zhonghong.ou@bupt.edu.cn, J.Cen@southampton.ac.uk\\
}

\usepackage{bibentry}
\begin{document}
\maketitle
\begin{figure*}[!ht]
    \centering
    \includegraphics[width=0.85\linewidth]{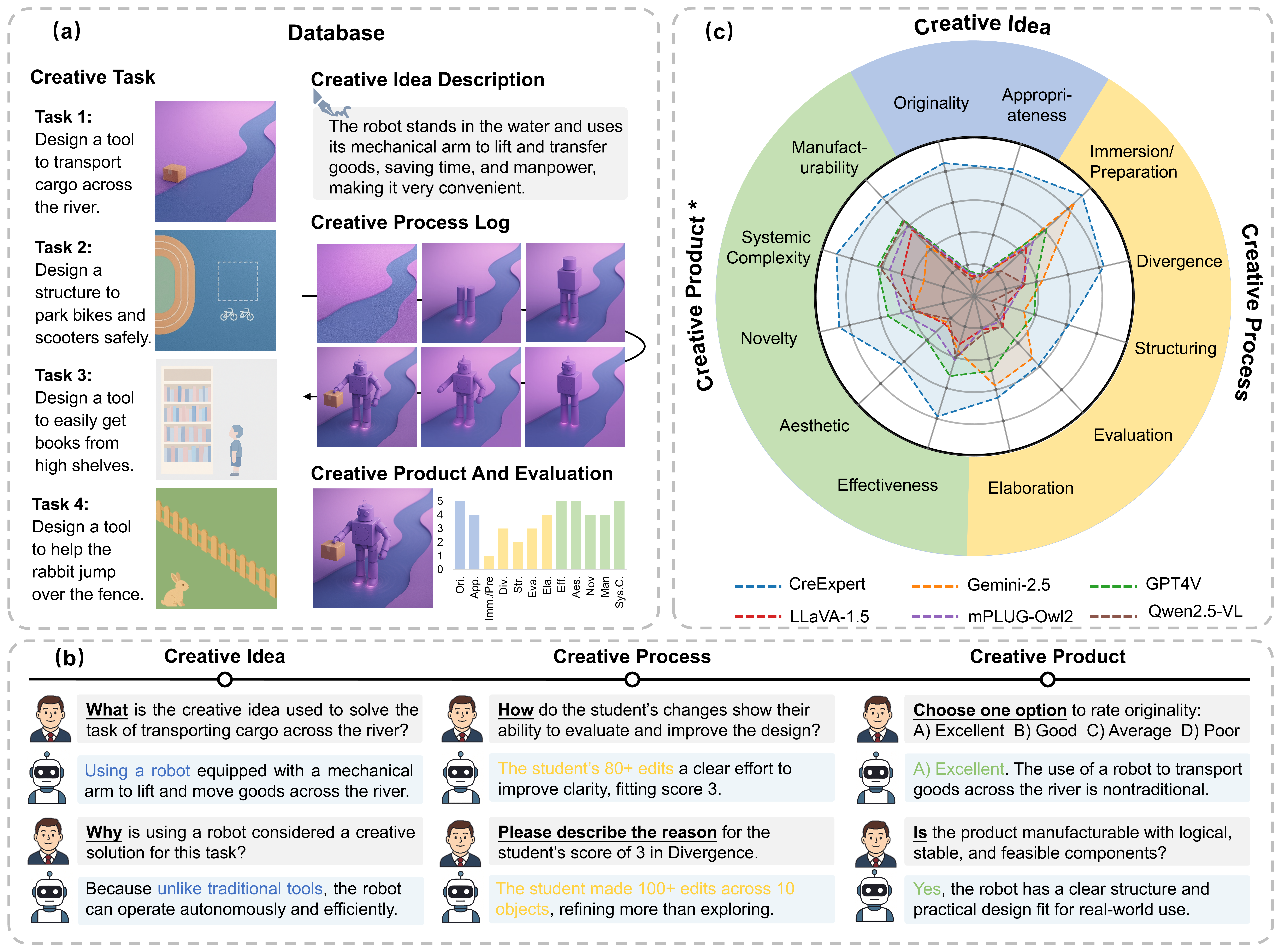}
    \vspace{-0.1in}
    \captionof{figure}{Overview of \textbf{CreBench}. (a) We design a diverse set of creative tasks and build a multi-dimensional database. (b) We use GPT-4o to generate instruction-following data through prompting. (c) Performance of the proposed \textbf{CreExpert} on various creativity evaluation dimensions. (*) indicates scaled data for better view.}
    \label{fig:teaser}
\end{figure*} 

\begin{abstract}
Human-defined creativity is highly abstract, posing a challenge for multimodal large language models (MLLMs) to comprehend and assess creativity that aligns with human judgments. The absence of an existing benchmark further exacerbates this dilemma. To this end, we propose \textbf{CreBench}, which consists of two key components: 1) an evaluation benchmark covering the multiple dimensions from creative idea to process to products; 2) \textbf{CreMIT} (\textbf{Cre}ativity \textbf{M}ultimodal \textbf{I}nstruction \textbf{T}uning dataset), a multimodal creativity evaluation dataset, consisting of 2.2K diverse-sourced multimodal data, 79.2K human feedbacks and 4.7M multi-typed instructions. Specifically, to ensure MLLMs can handle diverse creativity-related queries, we prompt GPT to refine these human feedbacks to activate stronger creativity assessment capabilities. CreBench serves as a foundation for building MLLMs that understand human-aligned creativity. Based on the CreBench, we fine-tune open-source general MLLMs, resulting in \textbf{CreExpert}, a multimodal creativity evaluation expert model. Extensive experiments demonstrate that the proposed CreExpert models achieve significantly better alignment with human creativity evaluation compared to state-of-the-art MLLMs, including the most advanced GPT-4V and Gemini-Pro-Vision.
\end{abstract}
\begin{links}
    \link{Project Page}{https://kaixuewen.github.io/Crebench}
\end{links}

\section{Introduction}

Creativity, a critical part of human cognition and intelligence, is usually seen as the ability to generate novel and valuable ideas, processes, or outcomes~\cite{creativemind}. Along with the remarkable advancements in MLLMs~\cite{gpt4v,llava} for image understanding, generation, and complex tasks reasoning, researchers are becoming concerned about whether these models have the potential to fully understand and assess human creativity. However, human-defined creativity is naturally highly abstract, subjective, and multidimensional. \textbf{Is the creativity that existing MLLMs understand what human want them to?} The answer is that the current MLLMs do not perform satisfactorily (as shown in Figure~\ref{fig:teaser}(c)). 

Despite the increasing demand for creativity-aware AI systems, evaluating creativity in MLLMs remains an underexplored area. Existing benchmarks for vision-language models mainly focus on tasks such as visual question answering~\cite{vqav2}, captioning~\cite{nocaps}, and image-text retrieval~\cite{flickr30k}, which are often objective and rely on well-defined ground truths. In contrast, creativity is inherently open-ended and lacks universally accepted evaluation criteria, making automatic assessment particularly challenging. Moreover, current evaluation metrics such as BLEU, CIDEr, or CLIPScore are insufficient for capturing the novelty, usefulness, and human-aligned imagination involved in creative tasks. Without reliable benchmarks or datasets tailored for creativity, it is difficult to measure how well MLLMs align with human notions of creativity or to guide their improvement in this dimension. 

To bridge this gap, we introduce \textbf{CreBench}, a comprehensive benchmark designed to evaluate MLLMs’ creativity alignment with human judgment across multiple dimensions from idea to process to product. The CreBench consists of two key parts: 1) To evaluate creativity in open-ended design tasks, we propose a comprehensive benchmark that captures creativity as a multi-stage process involving idea generation, iterative refinement, and expressive visual realization. Unlike traditional assessments that focus solely on final outcomes, our framework evaluates creativity across three core dimensions—creative idea, creative process, and creative product—each defined through twelve fine-grained indicators grounded in cognitive science, design reasoning, and creativity theory. A five-point behaviorally anchored rubric is used to assess each dimension, enabling precise, multidimensional analysis of students’ creative performance. By leveraging multimodal data—including participants’ verbalized ideas, interaction logs, and final visual outputs—this benchmark provides an ecologically valid foundation for studying how creativity unfolds in real-world problem-solving contexts. 2) To operationalize this framework and facilitate large-scale instruction tuning, we construct \textbf{CreMIT}, a high-quality multimodal dataset that captures the full spectrum of human and AI creativity. The dataset comprises over 2.2K creative instances and 79.2K expert-annotated evaluations, collected from four open-ended visual design tasks. All annotations are provided by trained experts following rigorous calibration protocols to ensure consistency and reliability. Furthermore, to scale instruction-level supervision for MLLMs, we use GPT-4o to transform expert feedback into 4.7M instruction–response pairs across six diverse QA formats (i.e., Reasoning, What, How, Why, Y/N, MCQ). CreMIT offers a unique resource for developing and benchmarking MLLMs in the domain of open-ended creativity assessment and reasoning.

To effectively assess creativity in real-world, open-ended problem-solving tasks, we introduce a structured evaluation framework that captures the multifaceted nature of creative thinking. Traditional approaches often focus narrowly on outcome novelty, overlooking the underlying cognitive and procedural dynamics. In contrast, CreBench integrates insights from creativity research, design cognition, and visual communication to provide a process-aware, multidimensional assessment. 

To evaluate performance on the CreBench, we propose \textbf{CreExpert}, a creativity evaluation expert model fine-tuned from open-source MLLMs. Built upon LLaVA-1.5~\cite{llava1.5}, CreExpert not only retain original knowledege, but also acquires human-aligned creativity understanding ablity. To the best of our knowledge, this is the first benchmark to systematically evaluate multimodal creativity.

In summary, our contributions are as follows:
\begin{itemize}
    \item We propose \textbf{CreBench}, comprising a fine-grained multi-dimensional creativity evaluation benchmark and a multimodal fine-tuning instruction dataset, named \textbf{CreMIT}. It contains 2.2K multi-source data, 79.2K individual human feedback, and 4.7M multi-typed instructions.
    \item We propose a multimodal creativity expert model, named \textbf{CreExpert}, built on CreBench via instruction fine-tuning. Extensive experiments demonstrate that the proposed CreExpert delivers significantly better creativity evaluation performances than the state-of-the-art MLLMs, including the most advanced GPT-4V and Gemini-Pro-Vision.
    \item \textbf{Open source} of our work release to the community, including: 1) A benchmark is used to evaluate the cretivity of MLLMs, and it consists of 12 fine-grain evaluation dimensions. 2) The multimodal instruction fine-tuning dataset CreMIT release in the community. 3) the proposed CreExpert, including codes and checkpoints.
\end{itemize}

\section{Related Works}
\subsection{Creativity in Image Generation}
Creativity, unlike aesthetics~\cite{aesbench,lapis}, emphasizes idea originality and imaginative depth rather than visual beauty alone. DALL-E~\cite{dalle}, and Stable Diffusion~\cite{stablediffusion} have significantly improved the quality and controllability of synthesized images. Several works have explored controlling generation style~\cite{styleclip}, enhancing compositionality~\cite{compositional}, or enabling prompt-based editing~\cite{instructpix2pix}. Creativity is often approximated via measures such as diversity or novelty, but these metrics oversimplify the complex human perception of creative value. In this work, we depart from this tradition by treating creativity as a multidimensional, subjective judgment and constructing a dataset explicitly labeled along 12 human-centric dimensions of creative perception.
% 不同于审美，创新性

\subsection{Human-Aligned Benchmark}
Human-aligned benchmark involves constructing evaluation datasets that reflect human preferences across various subjective dimensions. In language, benchmarks like HellaSwag~\cite{hellaswag}, TruthfulQA ~\cite{truthfulqa}, have been used to assess alignment with human reasoning and judgment. In multimodal domains, efforts such as Seed-Bench~\cite{seedbench} and MMBench~\cite{mmbench} extend this idea by integrating human preference into visual and cross-modal evaluation. For creative content, however, existing benchmarks either rely on pairwise preference~\cite{pick-a-pic} or aesthetic scoring~\cite{ava,laion-aesthetic}, failing to capture the nuanced, multidimensional structure of human creative judgment.
\subsection{Multimodal Large Language Model}
The release of LLMs, such as T5~\cite{T5}, GPT-4~\cite{gpt4}, and LLaMA~\cite{llama} in the community has attracted attention due to their remarkable performance. The success of LLMs has further motivated the multimodal community to explore vision-language interaction. Otter~\cite{otter} integrates visual perception and language instruction understanding, achieving task generalization through in-context instruction tuning. LLaVA~\cite{llava} introduces multimodal feature alignment and acquire multimodal reasoning capabilities via instruction tuning. Moreover, more works~\cite{detgpt,lisa,gpt4roi,deepseek-vl,llavanext} are devoted to using more fancy tricks to fine-tune MLLMs, in order to achieve better performance on downstream tasks. Despite these works making great progress, there is still a gap in terms of human-aligned creativity evaluation. To fill this gap, this work constructs the \textbf{CreBench} and \textbf{CreExpert} to fine-tune MLLM to understand human-aligned creativity.

\begin{figure*}[t]
    \centering
    \includegraphics[width=0.83\textwidth]{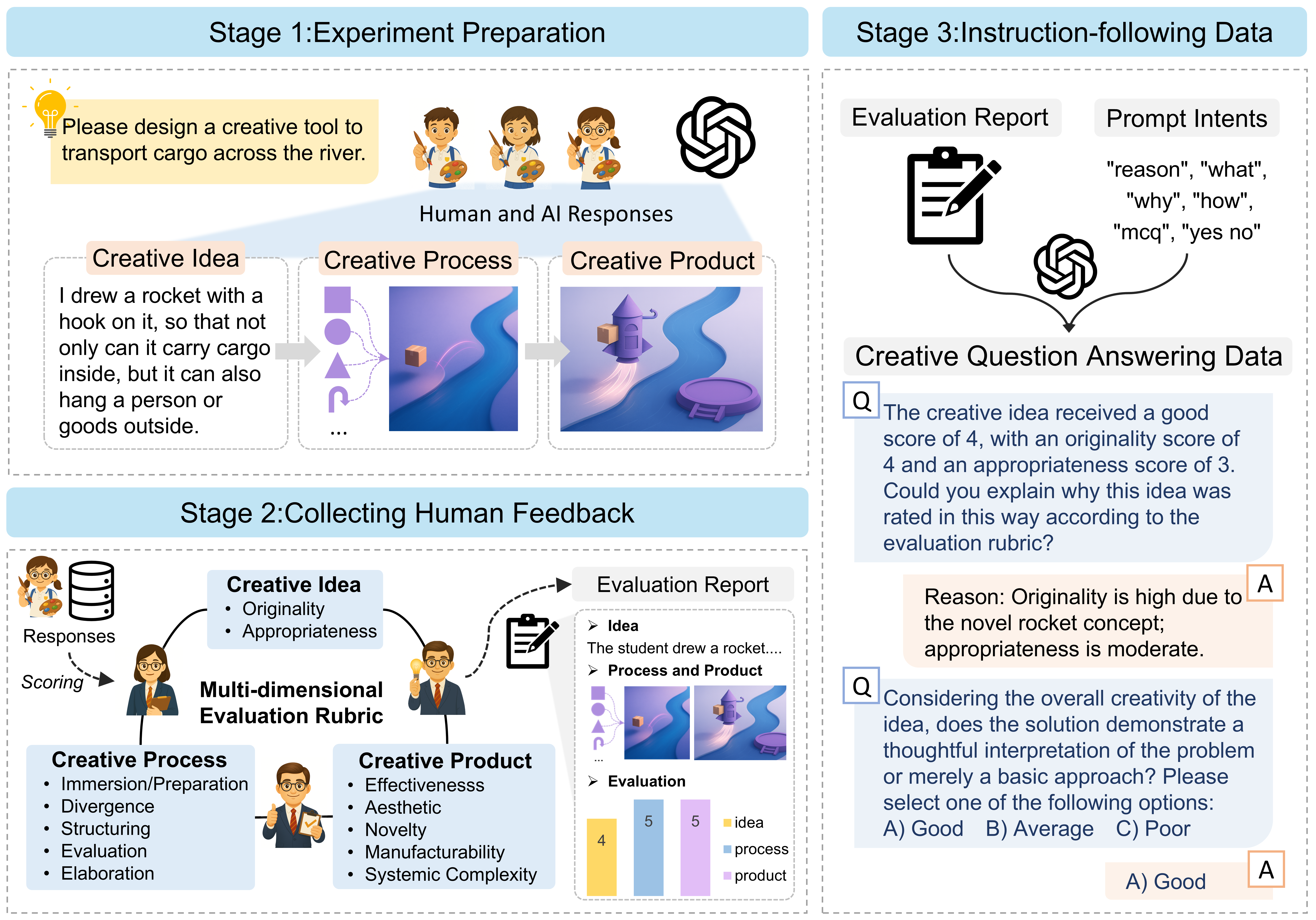}
    \vspace{-0.05in}
    \caption{Overview of \textbf{CreMIT} construction procedure. Stage 1: We collect diverse solutions (creative idea, creative process, and creative product) generated by students and AI based on open-ended creativity tasks. Stage 2: Innovation experts evaluate each solution across 12 indicators spanning three dimensions, producing detailed assessment reports. Stage 3:Using six types of prompts, we employ GPT-4o to generate multidimensional instruction-following data based on the expert feedback.}
    \vspace{-15pt}
    \label{fig:dataset}
\end{figure*}

\section{CreBench}

\begin{figure*}
    \centering
    \includegraphics[width=0.85\linewidth]{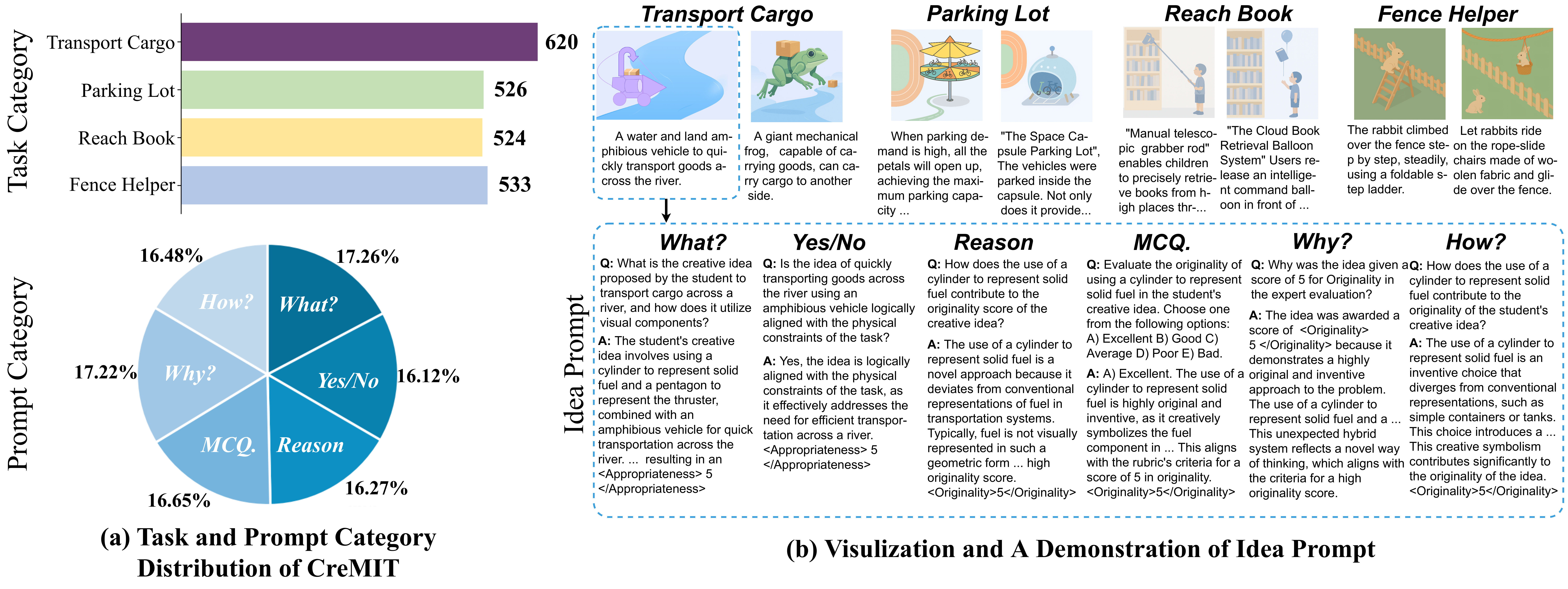}
    \vspace{-0.05in}
    \caption{\textbf{Data distribution and visualization.} (a) We analyze the sample number for each task and promt number of each prompt category. (b) We only demonstrate first sample of the idea prompt (among idea, process and product prompt).}%
    \vspace{-15pt}
    \label{fig:dataset_distribution}
\end{figure*}

\subsection{Evaluation Dimention Suite}
\label{sec:evaluation}
Assessing creativity in open-ended, real-world tasks demands a multidimensional framework beyond mere novelty. We adopt a comprehensive rubric tailored to creative problem solving and visual expression, dividing creativity into three dimensions: creative idea, creative process, and creative product (Figure~\ref{fig:dataset}). Creativity here involves generating, evaluating, and refining ideas \cite{brophy1998}, and manipulating visual elements \cite{urban2005assessing} to produce novel, appropriate, and aesthetically compelling solutions \cite{christensen2016dimensions}.
\subsubsection{Creative Idea}
The creative idea assesses the conceptual quality of a solution, focusing on its originality and appropriateness. It evaluates whether the idea presents novel yet contextually relevant mechanisms. Each aspect is rated on a five-point scale:

\begin{itemize}[leftmargin=*]
    \item \textbf{\textit{Originality}}: Measures the novelty and divergence from conventional approaches \cite{runco2012standard}.
\item \textbf{\textit{Appropriateness}}: Assesses the idea’s relevance, feasibility, and alignment with task requirements~\cite{judgments}.
\end{itemize}

\subsubsection{Creative Process}
This dimension measures students’ cognitive and visual engagement during problem solving, assessed across five aspects: immersion, divergence, structuring, evaluation, and elaboration. Each is rated on a five-point scale indicating the depth, coherence, and refinement of the creative process.

\begin{itemize}[leftmargin=*]
    \item \textbf{\textit{Immersion/Preparation}}: Initial engagement through reflection, observation, and strategic planning \cite{wallas1926art}.
\item \textbf{\textit{Divergence}}: Generation of varied and experimental ideas via open-ended exploration \cite{joy1950guilford}.
\item \textbf{\textit{Structuring}}: Intentional integration of visual elements into a coherent composition \cite{okada2017imitation}.
\item \textbf{\textit{Evaluation}}: Ongoing assessment and refinement of ideas to improve clarity and relevance \cite{mumford2013creative}.
\item \textbf{\textit{Elaboration}}: Attention to detail and expressive refinement in the final visual output \cite{urban2005assessing}.
\end{itemize}

\subsubsection{Creative Product}
This dimension evaluates the final drawing as both a solution and a creative expression, scored across five aspects: effectiveness, aesthetic, novelty, manufacturability, and systemic complexity. Each is rated on a five-point scale to assess the clarity, originality, feasibility, and design integration of the outcome.

\begin{itemize}[leftmargin=*]
    \item \textbf{\textit{Effectiveness}}: How clearly and coherently the drawing communicates the intended solution \cite{urban2005assessing}.
\item \textbf{\textit{Aesthetic}}: Visual appeal, composition balance, and expressive quality \cite{christensen2016dimensions}.
\item \textbf{\textit{Novelty}}: Originality in form, content, or symbolic representation \cite{torrance1966torrance}.
\item \textbf{\textit{Manufacturability}}: Feasibility of real-world construction and functionality \cite{charyton2011assessing}.
\item \textbf{\textit{Systemic Complexity}}: Integration of multiple functional components into a coherent system \cite{howard2008describing}.
\end{itemize}

\subsection{Dataset Construction}
This section presents a overview of the construction of the \textbf{CreMIT} dataset, as illustrated in Figure~\ref{fig:dataset}. Stage 1 outlines the preparation of a subjective experiment involving the collection of creative ideas, processes, and products from both human and AI participants across four distinct problem-solving tasks. Stage 2 covers the expert evaluation process, which yielded approximately 79.2K human feedback entries over 2.2K multi-dimensional evaluation instances. Stage 3 details how GPT-4o was prompted to refine creative reports and generate around 4.7M instruction-following samples across various creativity dimensions and question types.

\subsubsection{Experiment Preparation} Figure~\ref{fig:dataset} (Stage 1) illustrates the distribution of data by participant type, modality, and dimension.
\begin{enumerate}[label=\textbf{\protect\circled{\arabic*}}]
    \item \textbf{\textit{Task Design}}: We designed four open-ended, real-world problem-solving scenarios, such as the “cargo river crossing” (Figure~\ref{fig:dataset}, Stage 1), to elicit diverse and non-routine creative responses~\cite{runco2012divergent}. These tasks captured both the final creative products and the underlying ideas and processes, building a comprehensive, multimodal dataset for creativity assessment.

    \item \textbf{\textit{Subject Selection}}: A total of 512 secondary students were recruited through stratified cluster sampling from five schools to ensure demographic and cognitive diversity (Figure~\ref{fig:dataset}, Stage 1). All participants passed a color vision test~\cite{yang2022personalized} and completed a baseline creativity assessment. Each completed three creative tasks, contributing textual ideas, behavioral logs, and visual outputs—forming a multi-dimensional, multimodal dataset. This comprehensive data collection ensures a rich foundation for evaluating creativity across individuals with varied backgrounds and expression modalities.

\end{enumerate}

\begin{table}[!t]
\centering
\scriptsize
\setlength{\tabcolsep}{1.5mm}
\begin{tabular}{c|c|ccc|cccc}
\toprule
\multirow{2}{*}{Dataset} & \multirow{2}{*}{Year} & \multicolumn{3}{c|}{Modality} 
& \multirow{2}{*}{AIGC} 
& \multirow{2}{*}{\makecell{H.F.}} 
& \multirow{2}{*}{\makecell{I.T.}} 
& \multirow{2}{*}{\makecell{O.W.}} \\ 
&& Text & Proc. & Img & & & & \\ 
\midrule

OpenSketch & 2021 & \xmark & \xmark & \cmark & \xmark & \cmark & \xmark & \xmark \\ 

CID & 2022 & \cmark & \xmark & \cmark & \cmark & \cmark & \xmark & \cmark \\

Cambridge AUT & 2023 & \cmark & \xmark & \xmark & \xmark & \cmark & \xmark & \xmark \\ 

AesBench & 2024 & \cmark & \xmark & \cmark & \cmark & \cmark & \cmark & \cmark \\ 

APDD & 2024 & \xmark & \xmark & \cmark & \xmark & \cmark & \xmark & \xmark \\

IDEA & 2024 & \cmark & \xmark & \cmark & \xmark & \cmark & \xmark & \xmark \\
\textbf{CreBench} & 2025 & \cmark & \cmark & \cmark & \cmark & \cmark & \cmark & \cmark \\ 
\bottomrule
\end{tabular}
\vspace{-0.1in}
\caption{Feature comparison of creativity-related datasets. Proc. indicates process, i.e., creation log. H.F., I.T., and O.W. indicate human feedback, instruction tuning, open world, respectively.}
\label{tab:dataset}
\vspace{-0.3in}
\end{table}

\subsubsection{Collecting Human Feedback}
\begin{enumerate}[label=\textbf{\protect\circled{\arabic*}}]
    \item \textbf{\textit{Expert Annotation Protocol}}: To ensure reliable creativity assessment, we implemented a rigorous expert annotation and quality control protocol. Three experts in educational creativity were recruited following the Consensual Assessment Technique (CAT), each with extensive experience. Before annotation, all experts underwent two training sessions to familiarize themselves with the assessment framework, clarify rubrics, and calibrate standards using example cases—minimizing inter-rater variability. During annotation, experts evaluated both human and AI outputs across all three modalities. Annotation quality was ensured through ongoing agreement monitoring, regular calibration meetings, and both automated checks (e.g., completeness, consistency) and manual reviews for low-quality or ambiguous cases.
    
    \item \textbf{\textit{Expert Feedback}}: Based on this protocol, we constructed a multi-dimensional dataset through expert evaluations of 4 tasks, 3 creativity dimensions, and 12 sub-indicators. The dataset includes over 2.2K visual instances and 79.2K expert ratings. All annotations were conducted by qualified professionals to guarantee reliable creativity assessment. We have computed inter-rater reliability across all dimensions, the average Fleiss’s $\kappa$ = 0.71 and ICC (2,1) = 0.78, indicating substantial agreement.
\end{enumerate}

\subsubsection{Instruction-following Data} 
While the feedback dataset provides valuable knowledge for tuning MLLMs in creativity assessment, we further constructed an expanded instruction-following dataset to support diverse user queries and strengthen creativity perception. Following prior work~\cite{wu2024q}, we used GPT-4 to transform expert feedback into instruction–response pairs, as shown in Figure~\ref{fig:teaser}(b). This process yielded 4.7M instruction samples across multiple evaluation dimensions and question types.

To ensure MLLMs handle diverse queries in real-world settings, we designed six complementary question types:
\begin{enumerate}[label=\textbf{\protect\circled{\arabic*}}]
    \item \textbf{\textit{Reasoning-style}}: focus on analyzing the rationale behind expert-assigned scores, encouraging the model to justify the evaluation results based on rubric-defined criteria.
    \item \textbf{\textit{What-style}}: investigate the key features, intentions, or expressive elements of the creative idea that contribute to its originality and appropriateness;
    \item \textbf{\textit{How-style}}: explore implementation strategies or mechanisms, guiding the model to explain how the creative idea was realized or executed;
    \item \textbf{\textit{Why-style}}: seek to reveal the causes behind evaluation outcomes, encouraging causal interpretation;
    \item \textbf{\textit{Yes/No-style}}:questions provide binary judgments on aspects such as novelty, relevance, or feasibility, followed by brief justifications;
    \item \textbf{\textit{MCQ--style}}: convert evaluation scenarios into structured rating selections (Excellent to Bad), aligned with rubric-based judgment frameworks~\cite{llava1.5, lu2022learn, wu2024towards}.
\end{enumerate}

% idea
\begin{table*}[h]
\label{mllms}
\centering
\renewcommand{\arraystretch}{1.2}
\setlength{\tabcolsep}{1.5mm}
\resizebox{\textwidth}{!}{
\begin{tabular}{l|ccc|cccccc|cccccc|c|c}
\toprule
\multirow{2}{*}{\textbf{Model}}
& \multicolumn{3}{c|}{\textbf{Creative Idea}}
& \multicolumn{6}{c|}{\textbf{Creative Process}}
& \multicolumn{6}{c|}{\textbf{Creative Product}}
& \multirow{2}*{\textbf{Overall}}
& \multirow{2}*{\textbf{Rank}} \\
\cmidrule(lr){2-4}\cmidrule(lr){5-10}\cmidrule(lr){11-16}
& \emph{Ori.} & \emph{App.} & \emph{Avg.}
& \emph{Imm./Pre.} & \emph{Div.} & \emph{Str.} & \emph{Eva.} & \emph{Ela.} &  \emph{Avg.}
& \emph{Eff.} & \emph{Aes.} & \emph{Nov.} & \emph{Man.} & \emph{Sys. C.} &  \emph{Avg.}
& & \\
\midrule
% \rowcolor{Ocean!60}
\textbf{CreExpert (Ours)}
& \textbf{85.35\%} & \textbf{82.92\%} & \textbf{84.14\%}
& \textbf{92.24\%} & \textbf{82.97\%} & \textbf{61.64\%} & \textbf{58.97\%} & \textbf{65.13\%} & \textbf{72.19\%}
& \textbf{39.32\%} & \textbf{31.22\%} & \textbf{43.30\%} & \textbf{42.05\%} & \textbf{45.00\%} & \textbf{40.18\%}
& \textbf{65.50\%} & \textbf{1} \\
% \rowcolor{Ocean!40}
GPT-4V      
& 16.20\% & 14.12\% & 15.16\%
& 61.64\% & 39.48\% & 39.21\% & 36.70\% & 48.01\% & 45.01\%
& 26.10\% & 20.28\% & 27.64\% & 32.56\% & 31.63\% & 27.64\%
& 29.27\% & 2 \\
% \rowcolor{Ocean!20}
Gemini-Pro-Vision      
& 14.00\% & 8.93\% & 11.47\%
& 85.32\% & 43.72\% & 32.49\% & 53.07\% & 57.33\% & 54.39\%
& 17.75\% & 11.87\% & 19.84\% & 21.62\% & 16.41\% & 17.50\%
& 27.78\% & 3 \\
mPLUG-Owl2                 
& 15.16\% & 13.51\% & 14.34\%
& 50.29\% & 30.19\% & 23.05\% & 22.09\% & 20.91\% & 29.31\%
& 20.89\% & 16.46\% & 23.31\% & 30.46\% & 27.69\% & 23.76\%
& 22.47\% & 4 \\
LLaVA-1.5-7B           
& 13.23\% & 12.89\% & 13.06\%
& 44.11\% & 32.17\% & 20.02\% & 26.19\% & 21.41\% & 28.78\%
& 15.67\% & 12.63\% & 18.98\% & 28.35\% & 23.72\% & 19.87\%
& 20.57\% & 5 \\
InstructBLIP
& 9.26\% & 14.89\% & 12.08\%
& 46.92\% & 35.01\% & 19.42\% & 24.88\% & 22.77\% & 29.80\%
& 16.94\% & 9.83\% & 17.19\% & 29.34\% & 25.27\% & 19.71\%
& 20.53\% & 6 \\
Seed1.5-VL
& 19.97\% & 9.82\% & 14.90\%
& 57.74\% & 17.97\% & 11.65\% & 18.30\% & 18.26\% & 24.78\%
& 14.74\% & 26.27\% & 19.75\% & 21.03\% & 17.49\% & 19.86\%
& 19.85\% & 7 \\
Qwen2.5-VL
& 10.48\% & 14.23\% & 12.36\%
& 22.97\% & 32.70\% & 11.25\% & 25.62\% & 24.15\% & 23.34\%
& 19.84\% & 10.54\% & 20.12\% & 32.06\% & 30.74\% & 22.66\%
& 19.45\% & 8 \\
MiniGPT-v2
& 10.28\% & 11.89\% & 11.09\%
& 20.18\% & 30.83\% & 13.09\% & 22.10\% & 19.72\% & 21.18\%
& 18.00\% & 8.24\% & 14.90\% & 25.09\% & 21.27\% & 17.50\%
& 16.59\% & 9 \\
GLM
& 7.39\% & 9.19\% & 8.29\%
& 40.98\% & 28.31\% & 20.14\% & 20.58\% & 19.01\% & 25.80\%
& 12.09\% & 6.90\% & 10.03\% & 19.08\% & 17.93\% & 13.21\%
& 15.77\% & 10 \\
MiniGPT-4
& 5.89\% & 4.87\% & 5.38\%
& 10.23\% & 10.83\% & 9.42\% & 14.80\% & 10.08\% & 11.07\%
& 7.63\% & 10.23\% & 13.28\% & 6.72\% & 12.06\% & 9.98\%
& 8.81\% & 11 \\
TinyGPT
& 3.98\% & 2.59\% & 3.29\%
& 8.72\% & 4.80\% & 8.92\% & 10.17\% & 8.15\% & 8.15\%
& 7.22\% & 8.43\% & 9.82\% & 9.02\% & 4.95\% & 7.89\%
& 6.44\% & 12 \\
\bottomrule
\end{tabular}}
\vspace{-0.1in}
\caption{
Performance comparisons of the proposed \textbf{CreExpert} with existing MLLMs. Ori.: Originality, App.: Appropriateness, Imm./Pre.: Immersion/Preparation, Div.: Divergence, Str.: Structuring, Eva.: Evaluation, Ela.: Elaboration, Eff.: Effectiveness, Aes.: Aesthetic, Nov.: Novelty, Man.: Manufacturability, Sys. C.: Systemic Complexity. The Overall score is computed as the average across the three major dimensions: \textit{Creative Idea}, \textit{Creative Process}, and \textit{Creative Product}.
}
\vspace{-15pt}
\label{tab:main_result}
\end{table*}

Following this pipeline, the final \textbf{CreMIT} dataset comprises \textbf{4.7M multi-typed instructions} explicitly crafted to enhance creativity evaluation in MLLMs. As summarized in Table~\ref{tab:dataset}, CreMIT significantly advances in scale and diversity, offering a valuable benchmark for MLLM-based creativity understanding and assessment.

\section{CreExpert}
\subsection{Model Architecture}
The proposed \textbf{CreExpert} is developed following the architectural paradigm of LLaVA-1.5~\cite{llava}, and consists of three core modules: a vision encoder built upon CLIP-ViT-L14~\cite{clip} with a 336×336 input resolution, which encodes input images into 576 visual tokens; a modality bridging component implemented as a two-layer MLP that aligns visual and linguistic representations; and a language decoder initialized from the open-source Vicuna-v1.5~\cite{zheng2023judging}, which is responsible for instruction understanding and response generation. In this work, we instantiate the CreExpert model based on the LLaVA-1.5-7B variant, aiming to evaluate its capability in creativity assessment tasks using the \textbf{CreMIT} dataset.

\subsection{Supervised Fine-Tuning}
The development of open-source Multimodal Large Language Models (MLLMs)~\cite{llava} generally consists of two key phases: (1) aligning the visual encoder with the language model using large-scale image-text datasets collected from the web~\cite{lu2022learn}, and (2) enhancing multimodal reasoning through instruction tuning with curated vision-language datasets~\cite{llava}. In contrast to previous work that emphasizes general-purpose understanding~\cite{llava, pandagpt, mplug-owl}, the objective of this study is to improve the ability of MLLMs to perform creativity evaluation. To achieve this, we apply supervised instruction tuning to models that have been pre-trained on generic visual tasks, using the proposed \textbf{CreMIT} dataset. Drawing on insights from recent research~\cite{wu2024q}, our training method preserves the model’s general knowledge while equipping it with task-specific reasoning abilities tailored to creativity assessment. For the sake of training efficiency and fair comparison, the visual encoder is kept frozen, and fine-tuning is applied only to the projection module and the language model. As a result, we obtain a unified multimodal expert model that is capable of evaluating creativity across the full process, including idea generation, process execution, and final product realization.

\section{Experiments} 
\subsection{Dataset and Metrics} 
Our \textbf{CreBench} includes 2.2K samples sourced from 4 tasks, each sample contains a corresponding creative idea, creative process, and creative product. We split the dataset 50/50 into fine-tuning and evaluation sets. Besides, CreBench evaluates MLLMs via 12 fine-grained dimensions. We use the Pearson Correlation Coefficient as the evaluation metric, which reflects the consistency between the model's predictions and the actual human feedback, thereby expressing alignment with human judgment. 

\subsection{Implementation Details}
We fine-tuned pre-trained multimodal language models on the constructed \textbf{CreMIT} dataset using LoRA within the LLaMA-Factory framework. To ensure fair comparison, we adopted the default hyperparameter settings provided by the original models during the fine-tuning process. All training and evaluation were conducted on a server equipped with eight NVIDIA Tesla A40 48GB GPUs.

\subsection{Main Result}
We compare the performance of \textbf{CreExpert} with 11 top MLLMs. The comparison includes two widely used proprietary models, GPT-4V and Gemini Pro Vision, as well as 9 advanced open-source variants, i.e., LLaVA-1.5-7B, mPLUG-Owl2, InstructBLIP, Seed1.5-VL, Qwen2.5-VL, MiniGPT-v2, GLM, MiniGPT-4, and TinyGPT. All of these open-source models represent recent variants built upon foundational vision-language architectures. As shown in Table~\ref{tab:main_result}, CreExpert achieves the best performance, surpassing the state-of-the-art LLaVA-1.5-7B baseline by nearly 45\%. Among existing close-source models, GPT-4V~\cite{gpt4v} performs the best, yet still lags behind our \textbf{CreExpert} by more than 35\% in terms of overall score. These results demonstrate the superior creativity evaluation capability of \textbf{CreExpert} and highlight the effectiveness of the constructed \textbf{CreMIT} dataset in enhancing the alignment of multimodal foundation models.

\subsection{Ablation Study}

% idea
\begin{table}[!t]
\renewcommand{\arraystretch}{1.2}
\fontsize{7pt}{8pt}\selectfont
\centering
\setlength{\tabcolsep}{3.0mm}{
\begin{tabular}{c|c|cc|c}
\toprule
\multirow{2}{*}{\textbf{Task}} & \multirow{2}{*}{\textbf{Model}} 
& \multicolumn{2}{c|}{\textbf{Creative Idea}} & \multirow{2}{*}{\textbf{Overall}} \\
\cmidrule(lr){3-4}
&  & Ori. & App. & \\
\midrule
\multirow{3}{*}{Transport} 
& Baseline   & 12.80\% & 11.72\% & 12.26\% \\
& \textbf{CreExpert}  & \textbf{72.42\%} & \textbf{69.40\%} & \textbf{70.91\%} \\
& Improvement & +59.62\% & +57.68\% & +58.65\% \\
\bottomrule
\multirow{3}{*}{Parking} 
& Baseline   & 14.98\% & 14.96\% & 14.97\% \\
& \textbf{CreExpert}  & 69.08\% & 65.17\% & 67.13\% \\
& Improvement & +54.10\% & +50.21\% & +52.16\% \\
\bottomrule
\multirow{3}{*}{Reach} 
& Baseline   & 14.12\% & 15.06\% & 14.59\% \\
& \textbf{CreExpert}  & \textbf{83.91\%} & \textbf{78.07\%} & \textbf{80.99\%} \\
& Improvement & +69.79\% & +63.01\% & +66.40\% \\
\bottomrule
\multirow{3}{*}{Fence} 
& Baseline   & 11.90\% & 14.12\% & 13.01\% \\
& \textbf{CreExpert}  & \textbf{80.28\%} & \textbf{83.55\%} & \textbf{81.92\%} \\
& Improvement & +68.38\% & +69.43\% & +68.91\% \\
\bottomrule
\end{tabular}}
\vspace{-0.05in}
\caption{
Comparison of baseline MLLMs and the proposed \textbf{CreExpert} model on the \textbf{Creative Idea} dimension across four creative tasks.
}
\vspace{-15pt}
\label{tab:creative_idea}
\end{table}

\subsubsection{Creative Idea Evaluation Ability.}
From Table~\ref{tab:creative_idea}, we observe that fine-tuning baseline MLLMs with \textbf{CreMIT} significantly enhances their ability to generate creative ideas across various tasks. Among the four tasks evaluated, Transport, Parking, Reach, and Fence, the largest overall improvement is observed in the Fence task, where performance increased by almost 69\%. Notably, across all tasks, the Originality dimension consistently shows the most substantial gains, with improvements ranging from +54.10\% to +69.79\%. This suggests that MLLMs benefit particularly from \textbf{CreMIT} in generating novel ideas. A plausible explanation is that creative idea generation primarily involves textual expression, which enables MLLMs to better align with the key elements emphasized in human creativity evaluation, such as novelty and divergence.

% process
\begin{table}[!t]
\renewcommand{\arraystretch}{1.2}
\fontsize{7pt}{8pt}\selectfont
\centering
\setlength{\tabcolsep}{0.6mm}{
\begin{tabular}{c|c|ccccc|c}
\toprule
\multirow{2}{*}{\textbf{Task}} & \multirow{2}{*}{\textbf{Model}} 
& \multicolumn{5}{c|}{\textbf{Creative Process}} & \multirow{2}{*}{\textbf{Overall}} \\
\cmidrule(lr){3-7}
&  & Imm./Pre. & Div. & Str. & Eva. & Ela. \\
\midrule
\multirow{3}{*}{Transport} 
& Baseline   & 38.16\% & 25.87\% & 16.17\% & 22.31\% & 18.48\% & 24.20\% \\
& \textbf{CreExpert}  & \textbf{89.23\%} & \textbf{80.64\%} & \textbf{58.72\%} & \textbf{57.76\%} & \textbf{62.45\%} & \textbf{69.76\%} \\
& Improvement & +51.07\% & +54.77\% & +42.55\% & +35.45\% & +43.97\% & +45.56\% \\
\midrule
\multirow{3}{*}{Parking} 
& Baseline   & 41.53\% & 30.92\% & 20.25\% & 23.72\% & 21.56\% & 27.60\% \\
& \textbf{CreExpert} & \textbf{87.43\%} & \textbf{75.63\%} & \textbf{56.41\%} & \textbf{53.62\%} & \textbf{60.42\%} & \textbf{66.70\%} \\
& Improvement & +45.90\% & +44.71\% & +36.16\% & +29.90\% & +38.86\% & +39.10\% \\
\midrule
\multirow{3}{*}{Reach} 
& Baseline   & 39.07\% & 27.16\% & 28.32\% & 22.12\% & 19.53\% & 27.24\% \\
& \textbf{CreExpert}  & \textbf{90.52\%} & \textbf{81.83\%} & \textbf{59.73\%} & \textbf{56.49\%} & \textbf{61.27\%} & \textbf{69.97\%} \\
& Improvement & +51.45\% & +54.67\% & +31.41\% & +34.37\% & +41.74\% & +42.73\% \\
\midrule
\multirow{3}{*}{Fence} 
& Baseline   & 36.91\% & 25.47\% & 24.82\% & 19.16\% & 18.64\% & 25.00\% \\
& \textbf{CreExpert}  & \textbf{85.17\%} & \textbf{72.28\%} & \textbf{51.34\%} & \textbf{49.96\%} & \textbf{58.72\%} & \textbf{63.4\%} \\
& Improvement & +48.26\% & +46.81\% & +26.52\% & +30.8\% & +40.08\% & +38.49\% \\
\bottomrule
\end{tabular}}
\vspace{-0.1in}
\caption{
Comparison of baseline MLLMs and the proposed \textbf{CreExpert} model on the \textbf{Creative Process} dimension across four creative tasks.
}
\vspace{-15pt}
\label{tab:creative_process}
\end{table}

\subsubsection{Creative Process Evaluation Ability.}
Table~\ref{tab:creative_process} compares between the baseline MLLMs and the \textbf{CreExpert} on the Creative Process dimension across four creative tasks. The results demonstrate that \textbf{CreExpert} consistently outperforms the baseline across all sub-dimensions—including Immersion/Preparation, Divergence, Structuring, Evaluation, and Elaboration. The most substantial improvements are observed in Immersion/Preparation and Divergence, exceeding 50\% across most tasks. For example, in the Reach task, \textbf{CreExpert} improves Immersion/Preparation by +51.45\% and Divergence by +54.67\%. These results indicate that \textbf{CreMIT} enables MLLMs to evaluate novel ideas and simulate the multi-stage reasoning and exploration process underlying human creativity. Consistent gains in structuring and evaluation also suggest enhanced planning and critical thinking capabilities in the generated responses.

% product
\begin{table}[!t]
\renewcommand{\arraystretch}{1.2}
\fontsize{7pt}{8pt}\selectfont
\centering
\setlength{\tabcolsep}{0.5mm}{
\begin{tabular}{c|c|ccccc|c}
\toprule
\multirow{2}{*}{\textbf{Task}} & \multirow{2}{*}{\textbf{Model}} & \multicolumn{5}{c|}{\textbf{Creative Product}} & \multirow{2}{*}{\textbf{Overall}} \\
\cmidrule(lr){3-7}
& & Eff. & Aes. & Nov. & Man. & Sys. C. & \\
\midrule
\multirow{3}{*}{Transport} 
& Baseline & 13.35\% & 11.24\% & 17.54\% & 28.12\% & 22.69\% & 18.59\% \\
& \textbf{CreExpert} & \textbf{36.84\%} & \textbf{28.49\%} & \textbf{41.28\%} & \textbf{40.73\%} & \textbf{43.72\%} & \textbf{38.21\%} \\
& Improvement & +23.49\% & +17.25\% & +23.74\% & +12.61\% & +21.03\% & +19.62\% \\
\midrule
\multirow{3}{*}{Parking} 
& Baseline & 39.92\% & 24.81\% & 10.94\% & 20.67\% & 27.71\% & 24.81\% \\
& \textbf{CreExpert} & \textbf{53.45\%} & \textbf{30.17\%} & \textbf{20.26\%} & \textbf{28.89\%} & \textbf{39.84\%} & \textbf{34.52\%} \\
& Improvement & +13.53\% & +5.36\% & +9.32\% & +8.22\% & +12.13\% & +9.71\% \\
\midrule
\multirow{3}{*}{Reach} 
& Baseline & 22.94\% & 19.97\% & 17.28\% & 27.90\% & 26.13\% & 22.84\% \\
& \textbf{CreExpert} & \textbf{28.93\%} & \textbf{24.06\%} & \textbf{20.29\%} & \textbf{25.09\%} & \textbf{31.87\%} & \textbf{26.05\%} \\
& Improvement & +5.99\% & +4.09\% & +3.01\% & -2.81\% & +5.74\% & +3.21\% \\
\midrule
\multirow{3}{*}{Fence} 
& Baseline & 25.10\% & 11.09\% & 19.79\% & 19.18\% & 39.57\% & 22.95\% \\
& \textbf{CreExpert} & \textbf{30.35\%} & \textbf{28.62\%} & \textbf{35.86\%} & \textbf{32.16\%} & \textbf{48.76\%} & \textbf{35.15\%} \\
& Improvement & +5.25\% & +17.53\% & +16.07\% & +12.98\% & +9.19\% & +12.2\% \\
\bottomrule
\end{tabular}}
\vspace{-0.1in}
\caption{
Comparison of baseline MLLMs and the proposed \textbf{CreExpert} model on the \textbf{Creative Product} dimension across four creative tasks.
}
\vspace{-15pt}
\label{tab:creativity_product}
\end{table}
\subsubsection{Creative Product Evaluation Ability.}
Table~\ref{tab:creativity_product} compares the performance of baseline MLLMs and the \textbf{CreExpert} model on the Creative Product dimension across four tasks. \textbf{CreExpert} consistently improves over the baseline across all sub-dimensions, including Effectiveness, Aesthetics, Novelty, Manufacturability, and System Complexity. Notably, the Transport task exhibits the most prominent overall gain (+19.62\%), with improvements above 20\% in several key sub-dimensions such as Novelty (+23.74\%) and System Complexity (+21.03\%). In contrast, the Reach task shows only modest gains, with a slight drop in Manufacturability. These results suggest that while \textbf{CreMIT} strengthens the product-level creative expression of MLLMs, the degree of improvement varies across tasks depending on domain-specific constraints. Importantly, improvements in Novelty and Aesthetics—dimensions closely tied to user-perceived creativity—indicate that the model generates not only functional but also appealing and original solutions.

\section{Conclusion}
In this work, we attempt to leverage the innovation perception capability of multi-modality foundation models. Specifically, we first construct a corpus-rich, multi-dimensional innovation process evaluation database through human creative solutions, based on which we further establish a comprehensively annotated multimodal instruction tuning dataset for innovation processes (\textbf{CreMIT}). In addition, we propose multimodal innovation expert models based on innovation instruction fine-tuning, achieving significantly improved innovation perception performance. Building on these multimodal innovation expert models, we design a \textbf{CreExpert} to align model-based innovation evaluation with human judgments of innovation. We believe this work represents a solid step toward enhancing the innovation perception ability of MLLMs, and we hope that our contribution will inspire the research community to develop multi-modality foundation models capable of understanding highly abstract human creative processes.
\section{Acknowledgments}
This work is supported by the National Key Research and Development Program of China (Grant 2024YFC3308500), the Natural Science Foundation Joint Fund for Innovation and Development of Chongqing Municipal Education Commission (Grant CSTB2024NSCQ-LZX0132), the National Natural Science Foundation of China (Grant 62406036), the Beijing Municipal Natural Science Foundation ( Grant L251042), the State Key Laboratory of Networking and Switching Technology (Grant NST20250110), and the SMP-Zhipu.AI Large Model Cross-Disciplinary Fund (Grant ZPCG20241029322). Ethical approval for this study was obtained from the Ethics Committee of the University of Southampton, and written informed consent was obtained from all participants and their guardians.
\bibliography{aaai2026}

@article{gpt4v,
  title={The dawn of lmms: Preliminary explorations with gpt-4v (ision)},
  author={Yang, Zhengyuan and Li, Linjie and Lin, Kevin and Wang, Jianfeng and Lin, Chung-Ching and Liu, Zicheng and Wang, Lijuan},
  journal={arXiv preprint arXiv:2309.17421},
  volume={9},
  number={1},
  pages={1},
  year={2023}
}

@inproceedings{wu2024towards,
  title={Towards open-ended visual quality comparison},
  author={Wu, Haoning and Zhu, Hanwei and Zhang, Zicheng and Zhang, Erli and Chen, Chaofeng and Liao, Liang and Li, Chunyi and Wang, Annan and Sun, Wenxiu and Yan, Qiong and others},
  booktitle={European Conference on Computer Vision},
  pages={360--377},
  year={2024},
  organization={Springer}
}

@article{lu2022learn,
  title={Learn to explain: Multimodal reasoning via thought chains for science question answering},
  author={Lu, Pan and Mishra, Swaroop and Xia, Tanglin and Qiu, Liang and Chang, Kai-Wei and Zhu, Song-Chun and Tafjord, Oyvind and Clark, Peter and Kalyan, Ashwin},
  journal={Advances in Neural Information Processing Systems},
  volume={35},
  pages={2507--2521},
  year={2022}
}

@article{brophy1998,
  title={Understanding, measuring, and enhancing individual creative problem-solving efforts},
  author={Brophy, Dennis R},
  journal={Creativity Research Journal},
  volume={11},
  number={2},
  pages={123--150},
  year={1998},
  publisher={Taylor \& Francis}
}

@article{urban2005assessing,
  title={Assessing creativity: The Test for Creative Thinking-Drawing Production (TCT-DP).},
  author={Urban, Klaus K},
  journal={International Education Journal},
  volume={6},
  number={2},
  pages={272--280},
  year={2005},
  publisher={ERIC}
}

@article{christensen2016dimensions,
  title={Dimensions of creative evaluation: Distinct design and reasoning strategies for aesthetic, functional and originality judgments},
  author={Christensen, Bo T and Ball, Linden J},
  journal={Design studies},
  volume={45},
  pages={116--136},
  year={2016},
  publisher={Elsevier}
}

@article{runco2012standard,
  title={The standard definition of creativity},
  author={Runco, Mark A and Jaeger, Garrett J},
  journal={Creativity research journal},
  volume={24},
  number={1},
  pages={92--96},
  year={2012},
  publisher={Taylor \& Francis}
}

@article{judgments,
  title={Judgments of originality and appropriateness as predictors of creativity},
  author={Runco, Mark A and Charles, Robyn E},
  journal={Personality and individual differences},
  volume={15},
  number={5},
  pages={537--546},
  year={1993},
  publisher={Elsevier}
}

@book{wallas1926art,
  title={The art of thought},
  author={Wallas, Graham},
  number={24},
  year={1926},
  publisher={Harcourt, Brace}
}

@article{joy1950guilford,
  title={Guilford. 1950. Creativity},
  author={Joy, P},
  journal={American Psychologist},
  volume={5},
  number={9},
  pages={444--454},
  year={1950}
}

@article{okada2017imitation,
  title={Imitation, inspiration, and creation: Cognitive process of creative drawing by copying others' artworks},
  author={Okada, Takeshi and Ishibashi, Kentaro},
  journal={Cognitive science},
  volume={41},
  number={7},
  pages={1804--1837},
  year={2017},
  publisher={Wiley Online Library}
}

@article{torrance1966torrance,
  title={Torrance tests of creative thinking},
  author={Torrance, E Paul},
  journal={Educational and psychological measurement},
  year={1966}
}

@article{charyton2011assessing,
  title={Assessing creativity specific to engineering with the revised creative engineering design assessment},
  author={Charyton, Christine and Jagacinski, Richard J and Merrill, John A and Clifton, William and DeDios, Samantha},
  journal={Journal of Engineering Education},
  volume={100},
  number={4},
  pages={778--799},
  year={2011},
  publisher={Wiley Online Library}
}

@article{howard2008describing,
  title={Describing the creative design process by the integration of engineering design and cognitive psychology literature},
  author={Howard, Thomas J and Culley, Stephen J and Dekoninck, Elies},
  journal={Design studies},
  volume={29},
  number={2},
  pages={160--180},
  year={2008},
  publisher={Elsevier}
}

@article{aesbench,
  title={Aesbench: An expert benchmark for multimodal large language models on image aesthetics perception},
  author={Huang, Yipo and Yuan, Quan and Sheng, Xiangfei and Yang, Zhichao and Wu, Haoning and Chen, Pengfei and Yang, Yuzhe and Li, Leida and Lin, Weisi},
  journal={arXiv preprint arXiv:2401.08276},
  year={2024}
}

@inproceedings{lapis,
  title={LAPIS: A novel dataset for personalized image aesthetic assessment},
  author={Maerten, Anne-Sofie and Chen, Li-Wei and De Winter, Stefanie and Bossens, Christophe and Wagemans, Johan},
  booktitle={Proceedings of the Computer Vision and Pattern Recognition Conference},
  pages={6302--6311},
  year={2025}
}

@inproceedings{dalle,
  title={Zero-shot text-to-image generation},
  author={Ramesh, Aditya and Pavlov, Mikhail and Goh, Gabriel and Gray, Scott and Voss, Chelsea and Radford, Alec and Chen, Mark and Sutskever, Ilya},
  booktitle={International conference on machine learning},
  pages={8821--8831},
  year={2021},
  organization={Pmlr}
}

@inproceedings{stablediffusion,
  title={High-resolution image synthesis with latent diffusion models},
  author={Rombach, Robin and Blattmann, Andreas and Lorenz, Dominik and Esser, Patrick and Ommer, Bj{\"o}rn},
  booktitle={Proceedings of the IEEE/CVF conference on computer vision and pattern recognition},
  pages={10684--10695},
  year={2022}
}

@inproceedings{styleclip,
  title={Styleclip: Text-driven manipulation of stylegan imagery},
  author={Patashnik, Or and Wu, Zongze and Shechtman, Eli and Cohen-Or, Daniel and Lischinski, Dani},
  booktitle={Proceedings of the IEEE/CVF international conference on computer vision},
  pages={2085--2094},
  year={2021}
}

@inproceedings{compositional,
  title={Compositional visual generation with composable diffusion models},
  author={Liu, Nan and Li, Shuang and Du, Yilun and Torralba, Antonio and Tenenbaum, Joshua B},
  booktitle={European conference on computer vision},
  pages={423--439},
  year={2022},
  organization={Springer}
}

@inproceedings{instructpix2pix,
  title={Instructpix2pix: Learning to follow image editing instructions},
  author={Brooks, Tim and Holynski, Aleksander and Efros, Alexei A},
  booktitle={Proceedings of the IEEE/CVF conference on computer vision and pattern recognition},
  pages={18392--18402},
  year={2023}
}

@inproceedings{hellaswag,
  title={HellaSwag: Can a Machine Really Finish Your Sentence?},
  author={Zellers, Rowan and Holtzman, Ari and Bisk, Yonatan and Farhadi, Ali and Choi, Yejin},
  booktitle={Proceedings of the 57th Annual Meeting of the Association for Computational Linguistics},
  pages={4791--4800},
  year={2019}
}

@inproceedings{truthfulqa,
  title={TruthfulQA: Measuring How Models Mimic Human Falsehoods},
  author={Lin, Stephanie and Hilton, Jacob and Evans, Owain},
  booktitle={Proceedings of the 60th Annual Meeting of the Association for Computational Linguistics (Volume 1: Long Papers)},
  pages={3214--3252},
  year={2022}
}

@inproceedings{mmbench,
  title={Mmbench: Is your multi-modal model an all-around player?},
  author={Liu, Yuan and Duan, Haodong and Zhang, Yuanhan and Li, Bo and Zhang, Songyang and Zhao, Wangbo and Yuan, Yike and Wang, Jiaqi and He, Conghui and Liu, Ziwei and others},
  booktitle={European conference on computer vision},
  pages={216--233},
  year={2024},
  organization={Springer}
}

@inproceedings{seedbench,
  title={Seed-bench: Benchmarking multimodal large language models},
  author={Li, Bohao and Ge, Yuying and Ge, Yixiao and Wang, Guangzhi and Wang, Rui and Zhang, Ruimao and Shan, Ying},
  booktitle={Proceedings of the IEEE/CVF Conference on Computer Vision and Pattern Recognition},
  pages={13299--13308},
  year={2024}
}

@article{pick-a-pic,
  title={Pick-a-pic: An open dataset of user preferences for text-to-image generation},
  author={Kirstain, Yuval and Polyak, Adam and Singer, Uriel and Matiana, Shahbuland and Penna, Joe and Levy, Omer},
  journal={Advances in neural information processing systems},
  volume={36},
  pages={36652--36663},
  year={2023}
}

@inproceedings{ava,
  title={AVA: A large-scale database for aesthetic visual analysis},
  author={Murray, Naila and Marchesotti, Luca and Perronnin, Florent},
  booktitle={2012 IEEE conference on computer vision and pattern recognition},
  pages={2408--2415},
  year={2012},
  organization={IEEE}
}

@article{laion-aesthetic,
  title={Laion-5b: An open large-scale dataset for training next generation image-text models},
  author={Schuhmann, Christoph and Beaumont, Romain and Vencu, Richard and Gordon, Cade and Wightman, Ross and Cherti, Mehdi and Coombes, Theo and Katta, Aarush and Mullis, Clayton and Wortsman, Mitchell and others},
  journal={Advances in neural information processing systems},
  volume={35},
  pages={25278--25294},
  year={2022}
}

@inproceedings{wu2024q,
  title={Q-instruct: Improving low-level visual abilities for multi-modality foundation models},
  author={Wu, Haoning and Zhang, Zicheng and Zhang, Erli and Chen, Chaofeng and Liao, Liang and Wang, Annan and Xu, Kaixin and Li, Chunyi and Hou, Jingwen and Zhai, Guangtao and others},
  booktitle={Proceedings of the IEEE/CVF conference on computer vision and pattern recognition},
  pages={25490--25500},
  year={2024}
}

@article{llava,
  title={Visual instruction tuning},
  author={Liu, Haotian and Li, Chunyuan and Wu, Qingyang and Lee, Yong Jae},
  journal={Advances in neural information processing systems},
  volume={36},
  pages={34892--34916},
  year={2023}
}

@inproceedings{clip,
  title={Learning transferable visual models from natural language supervision},
  author={Radford, Alec and Kim, Jong Wook and Hallacy, Chris and Ramesh, Aditya and Goh, Gabriel and Agarwal, Sandhini and Sastry, Girish and Askell, Amanda and Mishkin, Pamela and Clark, Jack and others},
  booktitle={International conference on machine learning},
  pages={8748--8763},
  year={2021},
  organization={PmLR}
}

@article{zheng2023judging,
  title={Judging llm-as-a-judge with mt-bench and chatbot arena},
  author={Zheng, Lianmin and Chiang, Wei-Lin and Sheng, Ying and Zhuang, Siyuan and Wu, Zhanghao and Zhuang, Yonghao and Lin, Zi and Li, Zhuohan and Li, Dacheng and Xing, Eric and others},
  journal={Advances in neural information processing systems},
  volume={36},
  pages={46595--46623},
  year={2023}
}

@article{t5,
  title={Exploring the limits of transfer learning with a unified text-to-text transformer},
  author={Raffel, Colin and Shazeer, Noam and Roberts, Adam and Lee, Katherine and Narang, Sharan and Matena, Michael and Zhou, Yanqi and Li, Wei and Liu, Peter J},
  journal={Journal of machine learning research},
  volume={21},
  number={140},
  pages={1--67},
  year={2020}
}

@article{gpt4,
  title={Gpt-4 technical report},
  author={Achiam, Josh and Adler, Steven and Agarwal, Sandhini and Ahmad, Lama and Akkaya, Ilge and Aleman, Florencia Leoni and Almeida, Diogo and Altenschmidt, Janko and Altman, Sam and Anadkat, Shyamal and others},
  journal={arXiv preprint arXiv:2303.08774},
  year={2023}
}

@article{mplug-owl,
  title={mplug-owl: Modularization empowers large language models with multimodality},
  author={Ye, Qinghao and Xu, Haiyang and Xu, Guohai and Ye, Jiabo and Yan, Ming and Zhou, Yiyang and Wang, Junyang and Hu, Anwen and Shi, Pengcheng and Shi, Yaya and others},
  journal={arXiv preprint arXiv:2304.14178},
  year={2023}
}

@article{otter,
  title={Otter: A multi-modal model with in-context instruction tuning},
  author={Li, Bo and Zhang, Yuanhan and Chen, Liangyu and Wang, Jinghao and Pu, Fanyi and Cahyono, Joshua Adrian and Yang, Jingkang and Li, Chunyuan and Liu, Ziwei},
  journal={IEEE Transactions on Pattern Analysis and Machine Intelligence},
  year={2025},
  publisher={IEEE}
}

@article{detgpt,
  title={Detgpt: Detect what you need via reasoning},
  author={Pi, Renjie and Gao, Jiahui and Diao, Shizhe and Pan, Rui and Dong, Hanze and Zhang, Jipeng and Yao, Lewei and Han, Jianhua and Xu, Hang and Kong, Lingpeng and others},
  journal={arXiv preprint arXiv:2305.14167},
  year={2023}
}

@inproceedings{lisa,
  title={Lisa: Reasoning segmentation via large language model},
  author={Lai, Xin and Tian, Zhuotao and Chen, Yukang and Li, Yanwei and Yuan, Yuhui and Liu, Shu and Jia, Jiaya},
  booktitle={Proceedings of the IEEE/CVF Conference on Computer Vision and Pattern Recognition},
  pages={9579--9589},
  year={2024}
}

@inproceedings{gpt4roi,
  title={Gpt4roi: Instruction tuning large language model on region-of-interest},
  author={Zhang, Shilong and Sun, Peize and Chen, Shoufa and Xiao, Min and Shao, Wenqi and Zhang, Wenwei and Liu, Yu and Chen, Kai and Luo, Ping},
  booktitle={European conference on computer vision},
  pages={52--70},
  year={2024},
  organization={Springer}
}

@article{deepseek-vl,
  title={Deepseek-vl: towards real-world vision-language understanding},
  author={Lu, Haoyu and Liu, Wen and Zhang, Bo and Wang, Bingxuan and Dong, Kai and Liu, Bo and Sun, Jingxiang and Ren, Tongzheng and Li, Zhuoshu and Yang, Hao and others},
  journal={arXiv preprint arXiv:2403.05525},
  year={2024}
}

@misc{llavanext,
    title={LLaVA-NeXT: Improved reasoning, OCR, and world knowledge},
    url={https://llava-vl.github.io/blog/2024-01-30-llava-next/},
    author={Liu, Haotian and Li, Chunyuan and Li, Yuheng and Li, Bo and Zhang, Yuanhan and Shen, Sheng and Lee, Yong Jae},
    month={January},
    year={2024}
}

@inproceedings{yang2022personalized,
  title={Personalized image aesthetics assessment with rich attributes},
  author={Yang, Yuzhe and Xu, Liwu and Li, Leida and Qie, Nan and Li, Yaqian and Zhang, Peng and Guo, Yandong},
  booktitle={Proceedings of the IEEE/CVF Conference on Computer Vision and Pattern Recognition},
  pages={19861--19869},
  year={2022}
}

@article{runco2012divergent,
  title={Divergent thinking as an indicator of creative potential},
  author={Runco, Mark A and Acar, Selcuk},
  journal={Creativity research journal},
  volume={24},
  number={1},
  pages={66--75},
  year={2012},
  publisher={Taylor \& Francis}
}

@book{creativemind,
  title={The creative mind: Myths and mechanisms},
  author={Boden, Margaret A},
  year={2004},
  publisher={Routledge}
}

@inproceedings{vqav2,
  title={Making the v in vqa matter: Elevating the role of image understanding in visual question answering},
  author={Goyal, Yash and Khot, Tejas and Summers-Stay, Douglas and Batra, Dhruv and Parikh, Devi},
  booktitle={Proceedings of the IEEE conference on computer vision and pattern recognition},
  pages={6904--6913},
  year={2017}
}

@inproceedings{nocaps,
  title={Nocaps: Novel object captioning at scale},
  author={Agrawal, Harsh and Desai, Karan and Wang, Yufei and Chen, Xinlei and Jain, Rishabh and Johnson, Mark and Batra, Dhruv and Parikh, Devi and Lee, Stefan and Anderson, Peter},
  booktitle={Proceedings of the IEEE/CVF international conference on computer vision},
  pages={8948--8957},
  year={2019}
}

@inproceedings{flickr30k,
  title={Flickr30k entities: Collecting region-to-phrase correspondences for richer image-to-sentence models},
  author={Plummer, Bryan A and Wang, Liwei and Cervantes, Chris M and Caicedo, Juan C and Hockenmaier, Julia and Lazebnik, Svetlana},
  booktitle={Proceedings of the IEEE international conference on computer vision},
  pages={2641--2649},
  year={2015}
}

@article{llama,
  title={Llama: Open and efficient foundation language models},
  author={Touvron, Hugo and Lavril, Thibaut and Izacard, Gautier and Martinet, Xavier and Lachaux, Marie-Anne and Lacroix, Timoth{\'e}e and Rozi{\`e}re, Baptiste and Goyal, Naman and Hambro, Eric and Azhar, Faisal and others},
  journal={arXiv preprint arXiv:2302.13971},
  year={2023}
}

@article{pandagpt,
  title={Pandagpt: One model to instruction-follow them all},
  author={Su, Yixuan and Lan, Tian and Li, Huayang and Xu, Jialu and Wang, Yan and Cai, Deng},
  journal={arXiv preprint arXiv:2305.16355},
  year={2023}
}

@incollection{mumford2013creative,
  title={Creative thinking: Processes, strategies and knowledge},
  author={Mumford, Michael D and Giorgini, Vincent and Gibson, Carter and Mecca, Jensen},
  booktitle={Handbook of research on creativity},
  pages={249--264},
  year={2013},
  publisher={Edward Elgar Publishing}
}

@inproceedings{llava1.5,
  title={Improved baselines with visual instruction tuning},
  author={Liu, Haotian and Li, Chunyuan and Li, Yuheng and Lee, Yong Jae},
  booktitle={Proceedings of the IEEE/CVF conference on computer vision and pattern recognition},
  pages={26296--26306},
  year={2024}
}

\newpage
\clearpage
\onecolumn
\appendix
\section{Appendix}
\subsection*{Structure of the Assessment Framework}

The creativity assessment system used in this study distinguishes clearly between 
\textbf{assessment criteria} and \textbf{assessment rubrics}. The \textbf{criteria} 
specify \textit{what is being assessed}---that is, the conceptual dimensions and 
indicators that define creativity in this context. In contrast, the \textbf{rubrics} 
specify \textit{how the criteria are assessed}, operationalizing each indicator through 
a five-point performance scale. This distinction ensures conceptual clarity in the 
definition of creativity while providing a transparent and consistent scoring method 
for evaluating students' work.
\subsection{Creativity Assessment Criteria}

The assessment criteria define \textit{what aspects of creativity are evaluated} in this study. 
They outline the conceptual structure of the assessment framework by specifying the three core 
dimensions—\textbf{Creative Idea}, \textbf{Creative Process}, and \textbf{Creative Product}—along 
with their associated indicators. These criteria provide the foundation for interpreting students’ 
creative performance at a conceptual level, prior to the application of the analytic rubrics that 
operationalize each indicator through a five-point scale. The following subsections introduce the 
definition of creativity adopted in this study and present an overview of the criteria used to 
evaluate students’ drawings.
\subsubsection{Definition}

In this study, creativity is defined as the processes to generate, evaluate, and
improve problem-solving ideas, and to manipulate graphic elements to produce novel
and appropriate solutions. These processes involve divergent exploration of
possibilities, conceptual assembly of visual elements, refinement of core ideas, and
elaboration through detailed composition. The final drawing is expected to serve
both as a solution and as a creative expression, demonstrating originality,
aesthetic quality, and the ability to convey meaning in a visually impactful manner.
\subsubsection{Criteria Overview}
Table \ref{Criteria} presents the detailed innovation assessment criteria adopted in this study, providing a systematic and comprehensive evaluation of each work across three dimensions: creative ideation, creative processes, and creative products. For Creative Idea, the assessment emphasizes Originality and Appropriateness, examining whether the idea is novel, distinctive, and capable of effectively addressing the given problem. In the dimension of Creative Processes, the evaluation encompasses the full progression from Immersion/Preparation, Divergence, Structuring, and Evaluation to Elaboration, thereby capturing the depth and rigor demonstrated throughout the exploratory and developmental stages of creation. Finally, the evaluation of Creative Products focuses on Effectiveness, Aesthetic quality, Novelty, Manufacturability, and Systemic Complexity, which collectively determine the expressive quality and practical feasibility of the final output. Through this multi-layered set of criteria, the proposed assessment framework offers a holistic representation of the creative value and practical potential embodied in a work, from initial conceptualization to final realization.

\begin{table*}[h]
\centering
\footnotesize
\setlength{\tabcolsep}{6pt}
\renewcommand{\arraystretch}{1.25}

\begin{tabular}{p{5.3cm} p{5.3cm} p{5.3cm}}
\hline
\textbf{Creative Idea} 
& \textbf{Creative Process} 
& \textbf{Creative Product} \\
\hline

\textit{Originality} 
& \textit{Immersion / Preparation} 
& \textit{Effectiveness} \\
The idea is novel, surprising, or significantly different 
from conventional solutions.
&
Initial engagement with the task through reflection, 
observation, and preparation.
&
The drawing clearly communicates the intended solution or 
message in a relevant and coherent way.
\\
\hline

\textit{Appropriateness} 
& \textit{Divergence} 
& \textit{Aesthetic} \\
The idea is appropriate, relevant, and feasible in 
addressing the problem requirements.
&
The generation of varied and experimental ideas through 
open-ended and non-linear visual exploration.
&
The drawing is visually appealing, well-composed, and 
stylistically expressive.
\\
\hline

 & \textit{Structuring} & \textit{Novelty} \\
 & Integration of visual elements into a coherent, 
meaningful composition.
 &
The drawing reflects a novel or unconventional approach 
in content, form, or symbolism.
\\
\hline

 & \textit{Evaluation} & \textit{Manufacturability} \\
 & Assessment and adjustment of ideas to enhance clarity 
and relevance.
 &
The product can be realistically constructed or expected 
to function in real-world conditions.
\\
\hline

 & \textit{Elaboration} & \textit{Systemic Complexity} \\
 & Attention to detail and expressive refinement in 
the final product.
 &
The product consists of multiple functional components 
that interact as an integrated system to produce a 
coherent and purposeful solution.
\\
\hline
\end{tabular}

\caption{Criteria Overview: Creative Idea, Creative Process, and Creative Product}
\label{Criteria}
\end{table*}

\subsection{Creativity Assessment Rubrics}

Whereas the \textbf{assessment criteria} establish the conceptual dimensions and 
indicators used to characterize creativity, the \textbf{assessment rubrics} provide 
the analytic mechanism through which these criteria are evaluated. Each rubric 
operationalizes its corresponding indicators using a five-point performance scale, 
with descriptors that articulate gradations in quality, clarity, originality, and 
functional coherence. This approach ensures that the scoring process is both 
systematic and interpretable, allowing evaluators to identify meaningful differences 
in students’ creative performance across the dimensions of Creative Idea, Creative 
Process, and Creative Product. The detailed rubrics for each dimension are presented 
in the following subsections.

% creative idea
\subsubsection{Rubric for Evaluating the Creative Idea Dimension}
Table \ref{tab:creative_idea_rubric} presents a detailed five-point rubric for evaluating the “Creative Idea” dimension, specifically focusing on the two core criteria of Originality and Appropriateness. The rubric establishes clear descriptors for each performance level, ensuring consistency, transparency, and interpretability in scoring. For Originality, the scale ranges from highly inventive and conceptually unexpected designs (score 5) to generic solutions that replicate common ideas without novelty (score 1). For Appropriateness, the criteria span from ideas that fully and accurately address the problem context with a deep understanding of goals and constraints (score 5) to ideas that are misaligned with the task or fail to reflect its intent (score 1). Together, these descriptors provide a structured framework for evaluating the degree of innovation and contextual relevance demonstrated in students’ conceptual designs.

% creative process
\subsubsection{Rubric for Evaluating the Creative Process Dimension}
Table \ref{tab:creative_process_rubric} presents a five-point rubric for evaluating the “Creative Process” dimension, encompassing five essential criteria: Immersion/Preparation, Divergence, Structuring, Evaluation, and Elaboration. The rubric articulates clear behavioral descriptors for each performance level, capturing the depth and quality of students’ engagement throughout the creative process. These criteria span the full progression of creative activity—from initial orientation and exploratory breadth, to the organization of visual components, iterative refinement, and final expressive detailing. Scores range from highly intentional, exploratory, and well-structured design behaviors (score 5) to minimal preparation, limited experimentation, and low-refinement outputs (score 1). This framework enables a systematic assessment of students’ creative strategies and developmental pathways, offering a quantifiable means of understanding how their ideas evolve into visual solutions.

% creative product
\subsubsection{Rubric for Evaluating the Creative Product Dimension}
Table \ref{tab:creative_products_rubric} presents a five-point rubric for evaluating the “Creative Product” dimension, which includes five key criteria: Effectiveness, Aesthetic, Novelty, Manufacturability, and Systemic Complexity. These criteria collectively assess the overall quality, expressiveness, originality, feasibility, and structural integration of the final design output. The performance levels range from highly coherent, visually refined, inventive, and mechanically plausible solutions with strong system integration (score 5), to outcomes that are unclear, visually unorganized, conventional, structurally impossible, or lacking any functional coordination (score 1). This rubric provides a systematic means of evaluating the final products, enabling nuanced comparisons of how effectively different designs communicate ideas, achieve functional logic, and integrate multiple components into coherent systems.

% \caption{Rubric for Creative Idea (5-point scale).}
% \label{creative idea rubric}
% \end{table*}

% \subsubsection{creative idea}
\begin{table*}[h]
\centering
\footnotesize
\setlength{\tabcolsep}{4pt}
\renewcommand{\arraystretch}{1.25}

\begin{tabular}{p{2cm} p{3.0cm} p{2.8cm} p{2.8cm} p{2.8cm} p{2.8cm}}
\hline
\textbf{Creative Idea} 
& \centering\textbf{5} 
& \centering\textbf{4} 
& \centering\textbf{3} 
& \centering\textbf{2} 
& \centering\textbf{1} \\
\end{tabular}

\begin{tabular}{p{2cm} p{3.0cm} p{2.8cm} p{2.8cm} p{2.8cm} p{2.8cm}}
\hline
\textit{Originality} 
& Design is highly original, inventive, and conceptually surprising (e.g., an unexpected hybrid system).
& Design shows clear originality in structure, method, or symbolism.
& Design contains moderately new ideas or interesting combinations.
& Design shows minor variation but stays close to standard ideas.
& Design is generic, copying common bridge or boat ideas with no novel aspect.
\\
\hline

\textit{Appropriateness}
& The idea demonstrates a clear and accurate understanding of the problem context. It fully addresses the goal using a method that is logically aligned with the physical constraints and intended outcome. The solution reflects deep engagement with the task and shows that the student has thoughtfully interpreted what the problem demands.
& The idea is appropriate and context-aware. It aligns well with the problem's goals and constraints, showing a clear attempt to address the transportation challenge meaningfully.
& The idea generally fits the task. It reflects a basic understanding of the problem, but the approach may be underdeveloped or incomplete.
& The idea loosely relates to the task. It may address only a small aspect of the problem or show a superficial interpretation.
& The idea does not reflect an understanding of the task. It is irrelevant or completely misaligned with the goal.
\\
\hline

\end{tabular}

\caption{Rubric for Creative Idea (5-point scale).}
\label{tab:creative_idea_rubric}
\end{table*}

\begin{table*}[h]
\centering
\footnotesize
\setlength{\tabcolsep}{4pt}
\renewcommand{\arraystretch}{1.25}

\begin{tabular}{p{2.4cm} p{2.8cm} p{2.8cm} p{2.8cm} p{2.8cm} p{2.8cm}}
\hline
\textbf{Creative Process} 
&\centering\textbf{5}
&\centering\textbf{4}
&\centering\textbf{3}
&\centering\textbf{2}
&\centering\textbf{1}\\
\end{tabular}

\begin{tabular}{p{2.4cm} p{2.8cm} p{2.8cm} p{2.8cm} p{2.8cm} p{2.8cm}}
\hline
\textit{Immersion / Preparation} 
& Extended pause and active engagement with workspace before drawing; might include reading instructions, inspecting canvas, or naming layers.
& Noticeable preparation period with interface exploration.
& Brief period of exploration or canvas scanning before drawing; some interaction with tools.
& Very short delay before drawing; minimal or accidental interaction with tools or workspace.
& Begins drawing immediately without any delay; no signs of planning or understanding the task.
\\
\hline

\textit{Divergence}
& Extensive experimentation and visual exploration. The student tests a wide range of ideas and configurations, e.g., moving from linear to modular setups, trying floating, suspended, or submerged systems; switching object types and using unexpected arrangements. The behaviour is clearly open-ended, non-linear, and highly exploratory.
& Clear divergent thinking is visible. Multiple object types are explored (e.g., raft + crane + boat). Layouts and configurations are switched and re-evaluated. Actions like undoing and re-adding in new spatial zones suggest ongoing exploration.
& Moderate divergence is present. The student tests a few different objects and layouts; for example, placing ramps on both sides of the river, or comparing a bridge vs.\ a float. Exploration is present but somewhat cautious or short-lived.
& Minimal divergence is shown. The student tries one or two object types but keeps them in similar spatial zones or configurations. Exploration is hesitant and limited.
& The drawing follows a single idea or visual approach without variation. Only one object type is used, placed in a fixed position, with no signs of experimentation or change in direction.
\\
\hline

\textit{Structuring}
& The composition demonstrates a well-organized, intentional system. Components are spatially aligned and interlocked in a clear sequence. For example, cargo sits on a moving platform, pulled by a cable system, with all parts clearly contributing to the transport logic.
& Objects are purposefully arranged to reflect an intended mechanism. For example, a bridge is supported by pillars, cargo is placed on a barge, or ramps connect to functional ends. Functional pathways are mostly coherent.
& Basic structural logic begins to emerge. Components are loosely aligned or show directional consistency (e.g., ramps point toward water, platforms span across space). Visual order is forming, though incomplete.
& Some proximity or grouping is visible, but the relationships are unclear or accidental. Elements may be stacked without alignment, or misaligned in ways that obscure function.
& Objects are placed randomly with no visual or functional connection.
\\
\hline

\textit{Evaluation}
& Complex revision behaviour: deleting parts, replacing structures, or restructuring layout for better communication or accuracy.
& Multiple rounds of meaningful revision; clear improvement of design clarity or layout.
& Moderate adjustments like repositioning, resizing, or simple substitution.
& One or two minor adjustments with little effect on overall structure.
& No revision, deletion, or movement of objects; first attempt is final.
\\
\hline

\textit{Elaboration}
& Drawing demonstrates high refinement and expressive detail. Components are consistently scaled, aligned, and visually styled (e.g., contrast, rhythm, spacing). Layering, proportionality, and alignment all contribute to a polished and purposeful presentation.
& The design shows attention to visual clarity and composition. Components are well-aligned, proportionally adjusted, and refined with consistent colour choices or spacing. Some visual hierarchy is evident.
& Moderate refinement is shown. Objects are positioned thoughtfully, adjusted in size or shape, and some detail is applied. Layout begins to reflect visual intention.
& Some basic modifications are present such as resizing or flipping, but elements remain rough or simplistic. Limited use of visual features like alignment or layering.
& Drawing is minimal and lacks refinement. Only basic shapes are used with no visual differentiation or adjustments.
\\
\hline

\end{tabular}

\caption{Rubric for Creative Process (5-point scale).}
\label{tab:creative_process_rubric}
\end{table*}

\begin{table*}[h]
\centering
\footnotesize
\setlength{\tabcolsep}{4pt}
\renewcommand{\arraystretch}{1.25}

\begin{tabular}{p{2.4cm} p{2.8cm} p{2.8cm} p{2.8cm} p{2.8cm} p{2.8cm}}
\hline
\textbf{Creative Product} 
& \centering\textbf{5}
& \centering\textbf{4}
& \centering\textbf{3}
& \centering\textbf{2}
& \centering\textbf{1} \\
\end{tabular}

\begin{tabular}{p{2.4cm} p{2.8cm} p{2.8cm} p{2.8cm} p{2.8cm} p{2.8cm}}
\hline
\textit{Effectiveness}
& The drawing precisely and coherently communicates a well-developed solution. Every element contributes clearly to the intended function, and the viewer can readily follow how the system operates from start to finish.
& The drawing clearly presents the intended transport solution. Components are logically arranged, and their function in the overall system is understandable.
& The drawing generally communicates the solution. Key components are present (e.g., cargo, transport element, direction), though some relationships may be vague or incomplete.
& The drawing contains elements related to the task, but the relationship between components is unclear or fragmented. The viewer cannot infer how the system is meant to work.
& The drawing fails to communicate any coherent solution. It is either ambiguous or unrelated to the goal.
\\
\hline

\textit{Aesthetic}
& The drawing demonstrates high aesthetic refinement. All elements are harmoniously arranged with visual rhythm, balance, and consistency in scale and proportion. The style enhances the expressive impact of the solution.
& The composition is well-planned with clear attention to visual relationships. Spacing, alignment, and sizing are deliberate, and the layout supports clarity and coherence.
& The drawing shows moderate visual balance and organization. Objects are generally aligned, proportions are reasonable, and there is some consistency in size or spacing.
& Some attention to placement or shape is visible, but the composition lacks balance or coordination. Visual choices appear unintentional.
& The drawing appears cluttered, imbalanced, or visually inconsistent. Objects may overlap haphazardly, spacing is irregular, and there is no attention to layout or styling.
\\
\hline

\textit{Novelty}
& The drawing presents a highly creative and unconventional solution. It breaks away from typical patterns entirely, using imaginative content, structure, or symbolism that demonstrates unique and original thinking.
& The drawing shows a clearly original interpretation or representation. It uses inventive combinations, reimagined structures, or symbolic elements that differ significantly from common responses.
& The drawing includes one or two moderately novel aspects—such as an unexpected combination of objects or a personal visual metaphor—but still follows a recognizable structure.
& The drawing introduces slight variation on a common idea, such as minor changes in layout, shape, or visual elements, but the overall approach remains typical.
& The drawing uses a standard, highly conventional approach that closely resembles examples or common solutions. There is no noticeable deviation in concept, structure, or style.
\\
\hline

\textit{Manufacturability}
& The product is fully plausible and mechanically sound. All parts are logically positioned, structurally stable, and realistically constructible with attention to scale, support, and material.
& The product shows good awareness of physical logic and structural integrity. Components are well-supported and interconnected in ways that would likely work in practice.
& The product is generally feasible with minor structural issues. Most components appear buildable, but some simplification or assumptions would be required for it to function.
& The idea is partially grounded in reality but contains major structural or functional flaws (e.g., unclear connections, lack of stability). The design shows limited awareness of real-world constraints.
& The product includes elements that are structurally or physically impossible (e.g., gravity-defying structures). It cannot be realistically built under standard conditions.
\\
\hline

\textit{Systemic Complexity}
& The product demonstrates a fully integrated system of interdependent parts. Each component plays a distinct, purposeful role within a complex, layered structure, showing holistic design thinking.
& A well-coordinated system of components is present. Elements such as loaders, carriers, supports, and control units work together in a clear sequence to achieve the goal.
& The product includes multiple components with basic functional roles (e.g., cargo platform, bridge, pulley), and some connections or dependencies between them are evident.
& Two or more components are present, but they function independently or lack coordination. Their roles in the overall solution are unclear or loosely related.
& The product contains only one isolated component with no subsystems or interaction. There is no functional integration.
\\
\hline

\end{tabular}

\caption{Rubric for Creative Product (5-point scale).}
\label{tab:creative_products_rubric}
\end{table*}

\end{document}